\documentclass{article}
\usepackage{spconf}
\usepackage{bm}
\usepackage{epsfig}
\usepackage{graphicx}
\usepackage{subfigure}
\usepackage{float}
\usepackage{cite}
\usepackage{url}
\usepackage{color}
\usepackage{balance}
\usepackage{mdwlist}
\usepackage{multirow}
\usepackage{threeparttable}
\usepackage{enumitem}
\usepackage{amsmath}
\usepackage{amssymb}
\usepackage{stmaryrd}
\usepackage{booktabs}
\usepackage{siunitx}
\usepackage[numbers,sort&compress]{natbib}
\usepackage{epsfig,amsmath,amsfonts}
\usepackage{amsfonts}
\usepackage{verbatim}
\usepackage{amsthm}
\setenumerate[1]{itemsep=0pt,partopsep=0pt,parsep=\parskip,topsep=5pt}
\setitemize[1]{itemsep=0pt,partopsep=0pt,parsep=\parskip,topsep=5pt}

\makeatletter
\newif\if@restonecol
\makeatother

\usepackage[ruled,vlined]{algorithm2e}
\usepackage{algpseudocode}
\usepackage{amsmath}

\newcommand{\ie}{\textit{i}.\textit{e}.}
\newcommand{\eg}{\textit{e}.\textit{g}.}

\newcommand{\ten}[1]{\mathcal{#1}}
\newcommand{\mat}[1]{\mathbf{#1}}

\newcommand{\reff}[1]{(\ref{#1})}

\newtheorem{theorem}{Theorem}

\newtheorem{lemma}[theorem]{Lemma}

\newcommand{\argmax}{\mathop{\mathrm{argmax}}}

\newcommand{\st}{\mathop{\mathrm{s.t.}}}

\newcommand{\hzc}[1]{\textcolor{red}{#1}}

\begin{document}
\title{Active Sampling for Accelerated MRI with Low-Rank Tensors}

\name{Zichang He$^\star$, Bo Zhao$^\dagger$, and Zheng Zhang$^\star$}
\address{$^\star$ Department of Electrical and Computer Engineering, University of California, Santa Barbara\\ 
$^\dagger$Oden Institute for Computational Engineering and Sciences and\\ Department of Biomedical Engineering, University of Texas at Austin\\
Emails: zichanghe@ucsb.edu, bozhao@utexas.edu, zhengzhang@ece.ucsb.edu}


\maketitle
\begin{abstract}
Magnetic resonance imaging (MRI) is a powerful imaging modality that revolutionizes medicine and biology. 
The imaging speed of high-dimensional MRI is often limited, which constrains its practical utility. 
Recently, low-rank tensor models have been exploited to enable fast MR imaging with sparse sampling.
Most existing methods use some pre-defined sampling design, and active sensing has not been explored for low-rank tensor imaging. 
In this paper, we introduce an active low-rank tensor model for fast MR imaging.
We propose an active sampling method based on a Query-by-Committee model, making use of the benefits of low-rank tensor structure.
Numerical experiments on a 3-D MRI data set demonstrate the effectiveness of the proposed method. 
\end{abstract}

\section{Introduction}
\label{sec: intro}
Magnetic resonance imaging is a major medical imaging modality, which is widely used in clinical diagnosis and neuroscience. Due to the limited imaging speed, it is often highly desirable to speed up its imaging process. 
A lot of image models have been proposed to accelerate MR imaging, including sparsity-constrained~\cite{lustig2007sparse}, low-rank-constrained~\cite{liang2007spatiotemporal,lingala2011accelerated,zhao2012image}, data-driven, learning-based approaches, etc~\cite{ravishankar2019image}. 
Most of these methods recover MRI data with matrix computational techniques. 
They either focus on 2-D MRI problems or reshape the high-dimensional MRI data into a matrix and then solve the problem using matrix-based techniques. 

As a multi-dimensional generalization of matrix computation, tensor computation has been recently employed in MRI due to its capability of handling high-dimensional data~\cite{yu2014multidimensional,trzasko2013unified}. 
In many applications, MRI data sets naturally have a higher physical dimension. 
In these cases, tensors often better capture the hidden high-dimensional data pattern, achieving better reconstruction performance~\cite{he2016accelerated,kanatsoulis2019tensor}.

The quality and efficiency of an MRI reconstruction also highly depend on the sampling method.
Practical samples are measured in the spatial frequency domain of an MR image, often known as $k$-space.
Some adaptive sampling techniques have been proposed for matrix-format MR imaging based on compressive sensing or low-rank models~\cite{8632928,levine2017fly}. 
The experimental design methodologies include the Bayesian model~\cite{seeger2010optimization}, learning-based framework~\cite{gozcu2018learning,zhao2018optimal}, 
and so forth. Adaptive samplings may also be considered for streaming data~\cite{mardani2015subspace}. 

However, active sampling has not been explored for fast MR imaging with low-rank tensor models. 
Beyond the application of MR imaging, there are also limited works of adaptive sampling for tensor-structured data. 
Existing works mainly rely on the matrix coherence property~\cite{krishnamurthy2013low,liu2015adaptive}. 
They need to reshape the tensor to a matrix or can only apply on a 3-D tensor. 
Additionally, the practical MRI sampling is usually subject to some pattern constraints, like a Cartesian line sampling. 
Few papers have considered the pattern constraints caused by the practical MRI sampling.
Therefore, designing an active sampling method for tensor-structured data under certain pattern constraints is an important and open problem.

\textbf{Paper contributions.}
This paper presents an active sampling method for accelerating high-dimensional MR Imaging with low-rank tensors. 
Our specific contributions include:
\begin{itemize}[leftmargin=*]
    \item Novel active sampling methods for low-rank tensor-structured MRI data. A Query-by-Committee method is used to search for the most informative sample adaptively. Making use of the special tensor structure, the approximations towards the unfolding matrices naturally forms a committee. The sample quality is measured by the predictive variance, averaged leverage scores, or their combinations.
    \item Extending the sampling method to handle some pattern constraints in MR imaging. Our proposed sampling method itself can be applied broadly beyond MRI reconstruction.
    \item Numerical validation on an MRI example with Cartesian sampling. Numerical results show that the proposed methods outperform the existing tensor sampling methods.
\end{itemize}

\section{Background}
\label{sec:preliminaries}
\subsection{Notation}
Throughout this paper, a scalar is represented by a lowercase letter, \eg, $x$;
a vector or matrix is represented by a boldface lowercase or capital letter respectively, \eg, $\mat{x}$ and $\mat{X}$.
A tensor, which describes a multidimensional data array, is represented by a calligraphic letter. For instance, a $N$-dimensional tensor is denoted as $\ten{X} \in \mathbb{R}^{{I_1} \times {I_2}\ldots \times {I_N}}$, where ${I_n}$ is the mode size of the $n$-th mode (or dimension). 
An element indexed by $({i_1}, {i_2}\ldots, {i_N})$ in tensor $\ten{X}$ is denoted as $x_{{i_1}{i_2}\ldots{i_N}}$. 
A tensor Frobenius norm is defined as $\left\|\ten{X} \right\|_F := \sqrt{\sum\limits_{{i_1},{i_2},\ldots,{i_N}}({x_{{i_1}{i_2}\cdots{i_N}}})^2}$. 
A tensor $\ten{X} \in \mathbb{R}^{{I_1} \times {I_2}\cdots \times {I_N}}$ can be unfolded into a matrix along the $n$-th mode/dimension, denoted as ${\text{Unfold}_n}(\ten{X}):={\mat{X}_{(n)}} \in \mathbb{R}^{{I_n}\times {{I_1}\cdots{I_{n-1}}{I_{n+1}}\cdots{I_{N}} }}$.  
Conversely, folding the $n$-mode matrization back to the original tensor is denoted as ${\text{Fold}_n}({\mat{X}_{(n)}}):=\ten{X}$.

\subsection{Tensor-Format MRI Reconstruction}
In tensor completion, one aims to predict a whole tenser given only partial elements of the tensor, which is similar with matrix completion. 
In matrix cases, one often uses the nuclear norm ${\left\|  \cdot  \right\|_*}$ (sum of singular values) as a surrogate of a matrix rank, and seeks for a matrix with the minimal nuclear norm.
Exactly computing the tensor rank is NP-hard~\cite{hillar2013most}.
A popular heuristic surrogate of the tensor (multilinear) Tucker rank is the generalization of matrix nuclear norms~\cite{liu2012tensor}: 
\begin{equation}\label{eq:ten_nuclear}
{\left\| \ten{X} \right\|_*} = \sum\limits_{n = 1}^N {{{\left\| {{\mat{X}_{\left( n \right)}}} \right\|}_* }},
\end{equation}
Then the tensor completion problem can be formulated as minimizing Eq.~\reff{eq:ten_nuclear} given existing observations.

The above mode-$n$ matricization $\{{\mat{X}_{(n)}}\}^N_{n=1}$ represents the same set of data and thus are coupled to each other, which makes it hard to solve.
Therefore, we replace them with $n$ additional matrices $\{{\mat{X}_n}\}^N_{n=1}$ and introduce an additional tensor $\ten{M}$. The optimization problem can be reformulated as:
\begin{equation}\label{eq:raw_Obj}
\begin{aligned}
\mathop {\min }\limits_{\left\{\mat{X}_n\right\}^N_{n=1},\ten{M}} \quad& \sum\limits_{n = 1}^N {{{\left\| {{\mat{X}_n}} \right\|}_ * }}\\
\st \quad&{\mat{X}_ n} = {\mat{M}_{(n)}},\; n = 1,2,\ldots,N,\\
 & {\ten{M}_\Omega } = {\ten{T}_\Omega }, 
\end{aligned}
\end{equation}
where $\ten{M}$ is the reconstructed $k$-space tensor, $\mat{M}_{(n)}$ is the $n$-th mode matricization of $\ten{M}$, $\ten{T}$ is the fully sampled $k$-space tensor, and $\Omega$ is the observation set.
If the MRI data is known to hold some additional structure, we can further modify the imagining model, like a low-rank plus sparsity model~\cite{otazo2015low}. 
Eq.~\reff{eq:raw_Obj} can be efficiently solved via some alternating solvers, like the block coordinate descent and the alternating direction method of multipliers (ADMM)~\cite{liu2012tensor}. 
Essentially, we can either model the low-rankness of the $k$-space or the image space data.
We choose the former one since it enables the design of our active sampling methods in Section~\ref{sec:active_strategy}.
The additional $k$-space mode approximations $\left\{\mat{X}_n\right\}^N_{n=1}$ will be exploited to 
acquire new data.

\begin{figure}[!t]
    \centering
    \includegraphics[scale=0.38]{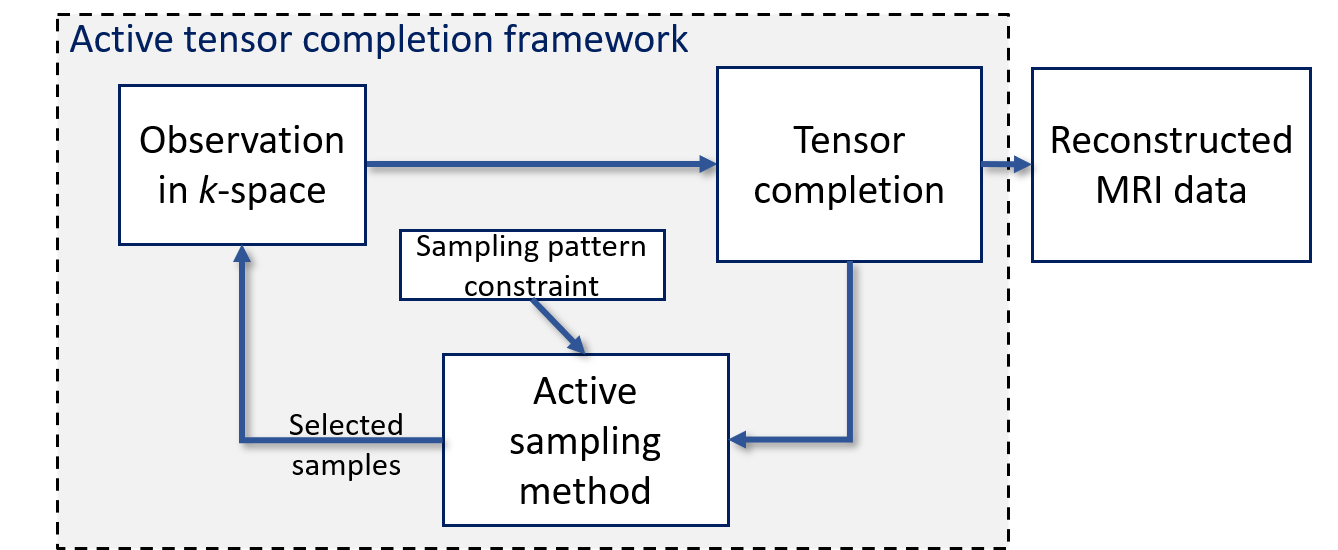}
    \caption{Flowchart of the MRI reconstruction framework.}
    \label{fig:whole_framework}
\end{figure}
\section{Active Tensor Sampling Method}
\label{sec:active_strategy}
The complete flowchart of our MRI reconstruction framework is illustrated in Fig.~\ref{fig:whole_framework}. In order to design an active sampling method for MRI reconstruction, two questions need to be answered: (1) How can we pick informative samples? (2) How can we guarantee that the new samples obey the patterns of MRI scans? We will provide the details in this section.

\subsection{Query-by-Committee-based Active Sampling}
\label{sec:element_sample}
We are inspired by a classical active learning approach called Query-by-Committee~\cite{seung1992query}. The key idea is to employ a committee of different models to predict the values at some candidate samples respectively. With such a committee, we can measure the quality of a candidate sample and pick the optimal one. The two key components of the Query-by-Committee approach are
\begin{itemize}[leftmargin=*]
    \item{\bf A committee of models.} The different mode-unfolding matrices obtained from solving Eq.~\reff{eq:raw_Obj} naturally form a committee required by our active sampling. 
    This model committee enables us to define an element-wise utility measure, denoted as $u(\xi)$, where $\xi$ is an element in the tensor. 
    \item {\bf Measure of sample quality.} After constructing a committee of models, we can define a measure of sample quality $u$. The sample with the maximized $u$ will be selected and added into the observation set: $\Omega \leftarrow \Omega \cup \argmax_{\xi}~u\left( {\xi} \right)$. We consider the predictive variance and leverage score as well as their combinations as our measure of sample quality, as detailed in the next sub-section.
\end{itemize} 

\subsection{Measure of Sample Quality}
{\bf Predictive variance.} The first choice is the predicted variance from different tensor modes. Based on the solved ${\mat{X}_n}$, if we unfold it and 
    enforcing its consistency with the observed data, we can obtain mode-$i$ low-rank approximation to $\ten{M}$:  
\begin{equation} 
\label{eq:mode_appro}
\begin{aligned}
& {\tilde{\ten{M}}_n} {\rm{ := }}  {{{\text{Fold}_n}(\mat{X}}_n}), \quad {\tilde{\ten{M}}_n}(\Omega) \leftarrow  \ten{T}(\Omega).
\end{aligned}
\end{equation}
Let $\mathbb{E}\left[ {\tilde{\ten{M}}} \right] := \sum_{n = 1}^N {{w_n}{{\tilde{\ten{M}}}_n}}$ be the reconstructed tensor,
we define the difference tensor of each approximation ${\Delta \ten{M}_n}$ and the predictive variance tensor $\ten{V}$ as:
\begin{equation}
\label{eq:var_measure}
{\Delta \ten{M}_n} := {\tilde{\ten{M}}_n} - \mathbb{E}\left[ {\tilde{\ten{M}}} \right],\; \; \ten{V} := \sum\limits_{n = 1}^N {{w_n}} {\left( {\Delta \ten{M}_n} \circ {\Delta \ten{M}_n} \right)}
\end{equation}
where $\circ$ denotes a Hadamard product, and ${w_n}$ is the weight associated with mode-$n$ approximation ${{\tilde{\ten{M}}}_n}$. The value of ${w_n}$ depends on the solver to Eq.~\reff{eq:raw_Obj}. In our implementation, we adopt an ADMM solver and have $\{w_n\}_{n=1}^N = \tfrac{1}{N}$.

Note that the predicted variance is a semi-variance that is defined to measure the disagreement among different modes. Since the consistency of the observed data is enforced, the committee has zero disagreement on them. Maximizing the predictive variance helps to identify the sample with large prediction error.
The following lemma describes the relation between the reconstructed error and the predictive variance.
\begin{lemma}
Let ${\mu}$ be the sum of all elements of the predictive variance tensor
$\mu:= \|{\rm vec}(\ten{V}) \|_1 = \sum\limits_{n = 1}^N {{w_n}} {\left\| {\Delta \ten{M}_n} \right\|_{\rm F}^2}$, ${\mu_i}$ be the mode-$n$ approximation error: ${\mu_n}=\left\|\ten{M} - {\tilde{\ten{M}_n}} \right\|_{\rm F}^2$, and ${\mu _{\rm rec}}$ be the reconstruction error of the whole data set: $\mu _{\rm rec}=\left\|\ten{M}-\mathbb{E}\left[ {\tilde{\ten{M}}} \right] \right\|_{\rm F}^2$, we have
\begin{equation}
\label{eq:relation_err_var}
{\mu _{\rm rec}} = \sum\limits_{n = 1}^N {w_n}{\mu_n } - \mu.
\end{equation}
\end{lemma}
The approximations of different modes gradually converge to the same one as more samples are observed. 
A sample with the maximized predictive variance help to $\mu_{\rm rec}$ decreases quickly. The similar idea is used to reduce generalization error in the ensemble  learning~\cite{mendes2012ensemble}.

{\bf Leverage score.} Our second quality measurement $u$ is the leverage score. It comes from the incoherence property of a matrix~\cite{candes2009exact}.
Given the singular value decomposition (SVD) of a rank-$R$ matrix $\mat{A}= \mat{U}\mat{\Sigma} {\mat{V}^H} \in {\mathbb{C}^{{N_1}\times{N_2}}}$, let $\mat{e}_i$ be the $i$-th standard basis, 
the left and right leverage scores of a matrix are defined as:
\begin{equation}
\begin{array}{l}
{\boldsymbol{\ell}(i)} := \frac{N_1}{R}{\left\| {{\mat{U}^T}{\mat{e}_i}} \right\|^2_2},\quad i = 1,2,\ldots,{N_1}, \\
{\mat{r}(j)} := \frac{N_2}{R}{\left\| {{\mat{V}^T}{\mat{e}_j}} \right\|^2_2},\quad j = 1,2,\ldots,{N_2}.
\end{array}
\end{equation}
A leverage score measures the coherence of a row/column with a coordinate direction, which generally reflect the importance of the rows/columns.
For the mode-$n$ approximation $\mat{X}_n$, we can perform SVD and calculate its left and right leverage scores as $\boldsymbol{\ell}_n$ and $\mat{r}_n$ respectively, then the element-level leverage score of a sample $\left(i,j \right)$ in $\mat{X}_n$ is defined as:
\begin{equation}
\label{eq:mode_lev_measure}
{{l_n} \left(i,j\right)}: = {\boldsymbol{\ell}_n(i)} \cdot {\mat{r}_n(j)},\quad n=1,2,\ldots,N.
\end{equation}
In a tensor structure, we can average the element-level leverage scores over modes and treat it as the utility measurement:
\begin{equation}
\label{eq:lev_measure}
\ten{L} := \sum\limits_{n = 1}^N {w_n} {\text{Fold}_n}\left( {\mat{L}_n} \right).
\end{equation}

Based on the above two measurements, we propose four greedy methods for our active tensor sampling as follows. Since the sampling performance is problem-dependent, we can not conclude the best one for an arbitrary data set. But they all perform well in our numerical experiments.
\begin{itemize}[leftmargin=*]

\item \textbf{Method 1 (Var):}  
We take the utility measurement $u$ as the variance among different modes [Eq.~\reff{eq:var_measure}]. 

\item \textbf{Method 2 (Lev):} 
We take the utility measurement $u$ as the (weighted) average leverage scores [Eq.~\reff{eq:lev_measure}]. 

\item \textbf{Method 3 / 4 (Var + Lev/ Var $\times$ Lev):} We take the utility measurement $u$ as the sum/ Hadamard product of the normalized variance [Eq.~\reff{eq:var_measure}] and the leverage scores [Eq.~\reff{eq:lev_measure}]. 


\end{itemize}
\begin{figure*}[t]
    \centering
    \includegraphics[width=6.6in]{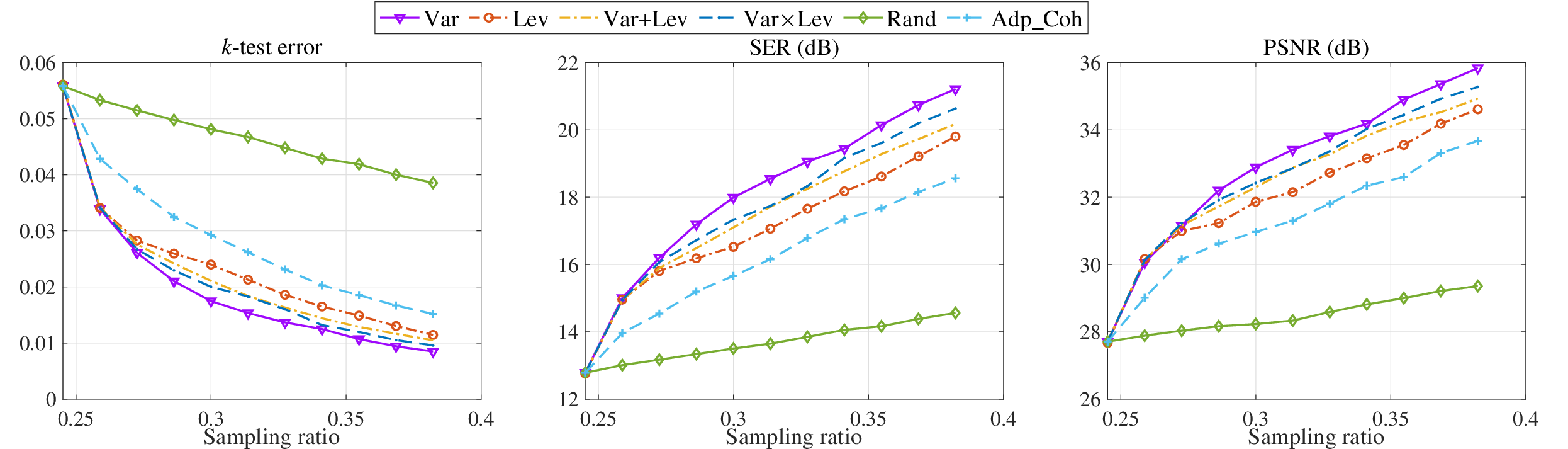}
    \caption{Reconstruction results of different adaptive sampling methods. The proposed methods all outperform the existing Ada\_Coh and random methods. Method 1 (Var) is the most suitable one in this example.} 
    \label{fig:combine_act}
\end{figure*}

\subsection{Sampling under Pattern Constraint}
\label{sec:pattern_sample}
Now we show how to pick the samples subject to some pattern constraints. 
Suppose the unobserved $k$-space data is partitioned into $P$ patterns $\{\Pi_i\}_{i=1}^P$, 
we should sample a certain pattern $\Pi_i$ rather than just one element.
For example, in a Cartesian sampling, the pattern ${\Pi_i}$ is a rectilinear sampling, \ie~a full column or row in a $k$-space matrix (${k_x}$-${k_y}$).

Since our utility measurements are calculated element-wisely in the Query-by-Committee, they can be easily applied to any sampling pattern by summing the utility measurement over all elements included in the pattern. 
Consequently, we have a pattern-wise measurement:
\begin{equation}\label{eq:pattern_calculate}
u\left( \Pi_i \right) = \begin{matrix} \sum_{\xi \in \Pi_i} {u\left( \xi \right)} \end{matrix}, \quad i = 1, 2, \ldots, P.
\end{equation}
As a result, the observation set can be updated as $\Omega \leftarrow \Omega \cup \argmax_{\Pi_i}~u\left( {\Pi_i} \right) $. 
Without any pattern constraints, Eq,~\reff{eq:pattern_calculate} will degenerate to an element-wise measurement. 

In practical implementations, the active learning algorithm can be easily extended to a batch version via selecting samples with the top-$K$ utility measurement. We can stop the algorithm when the algorithm runs out of a sampling budget. Note that beyond MRI reconstruction, the proposed active learning method is suitable for various tensor completion applications with different pattern constraints. The only requirement is that the low-rank tensor completion algorithms model the (multilinear) Tucker rank.

\section{Numerical results}
\label{sec:numerical_results}
In this section, we validate our active sampling methods on a low-rank tensor-format brain data set with the size $256 \times 256 \times 256$. Our codes are implemented in MATLAB and run on a computer with 2.3GHz CPU and 16GB memory. 
\begin{figure}[t]
    \centering
    \includegraphics[width=1\columnwidth]{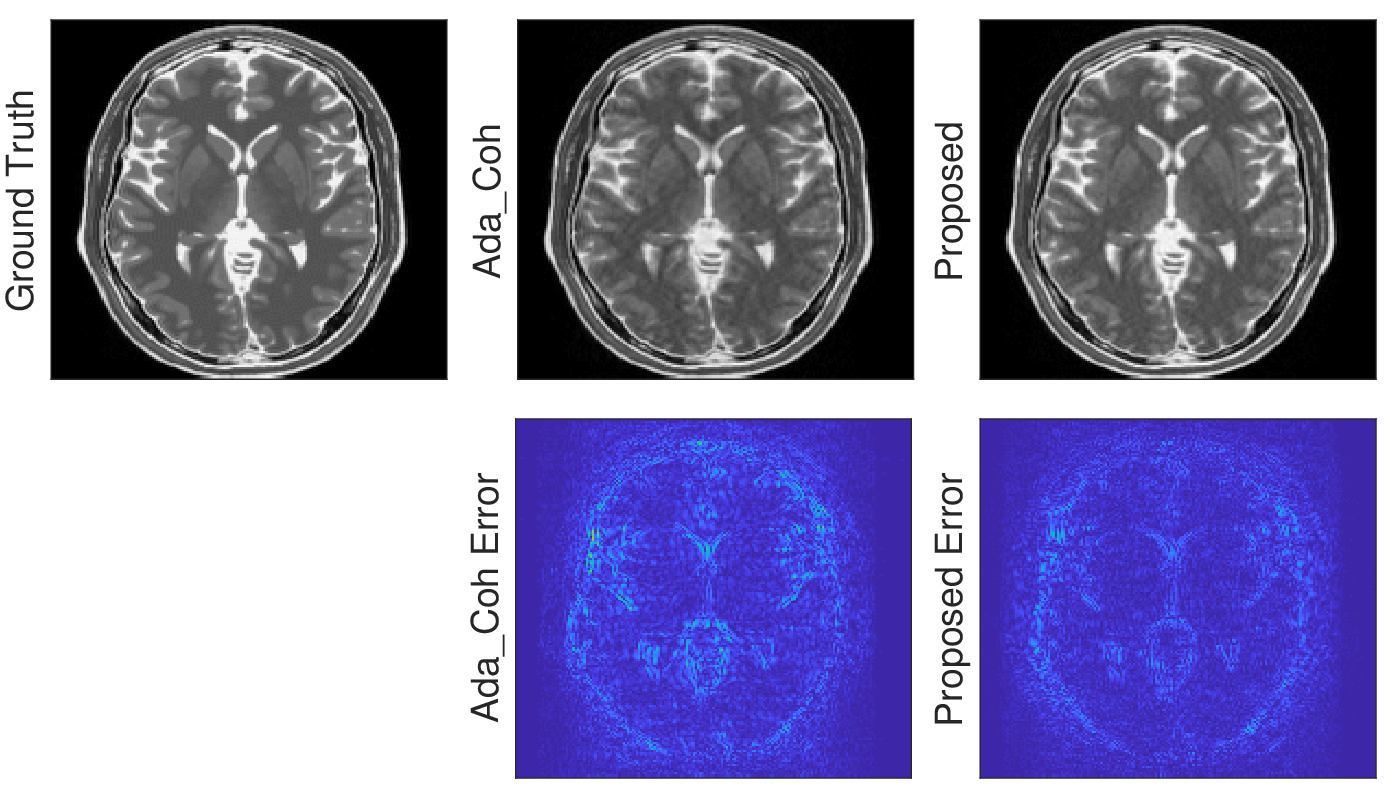}
    \caption{One slice of the reconstructed data.}
    \label{fig:Spatial}
\end{figure}


{\bf Baselines for Comparisons.} 
We reconstruct the tensor-structured data via adopting the proposed active learning methods, a matrix-coherence-based adaptive tensor sampling method (denoted as Ada\_Coh)~\cite{krishnamurthy2013low}, and a random sampling method. In Ada\_Coh~\cite{krishnamurthy2013low}, the sampling is driven by the coherence of one mode matricization of a tensor. Our Method 2 (Lev) can be seen as its generalization via taking consideration of the interactions among different modes. 

\textbf{Sampling Patterns.} In this example, the $k$-space data is scanned continuously as a fiber in the Cartesian coordinate. In each spatial slice, we initialize with a fully sampled center and randomly sampled non-central regions. The sample pattern of the candidate samples is a fiber in the $k$-space, which is obtained via fixing all but one index of the tensor.

\textbf{Evaluation metrics.}
Our methods sample and predict $k$-space data, and the final reconstruction needs to be visualized in the image space. Therefore, we choose the evaluation metrics from both spaces. 
In the $k$-space, the accuracy is measured by the relative mean square error evaluated over the fully-sampled $k$-space data (denoted as $k$-test). 
In the image space, we use the signal to error ratio (SER) and peak signal to noise ratio (PSNR) as our metrics.

\textbf{Results Summmary.} The initial sampling mask is generated with a sampling ratio of 23.14\%, including a 14.45\% fully sampled center. In active sampling, each sampling batch includes 1000 fibers, and we select a total of 10 batches sequentially. 
Fig.~\ref{fig:combine_act} plots the $k$-test, SER and PSNR as the sampling ratio increases. 
All proposed active sampling methods can significantly improve the evaluation metrics and outperform the existing Ada\_Coh and random samplings. We choose Method 1 (Var) for the further comparison. Fig.~\ref{fig:Spatial} shows one slice of the reconstructed MRI data. The evaluations of the whole data set are shown in Table~\ref{tab:evaluation}, where the proposed method has a higher reconstruction accuracy.
\begin{table}[t]
\centering
\caption{Evaluations on the whole data set.}
\label{tab:evaluation}
\begin{tabular}{cccc}
\toprule
 Sampling method    & $k$-test (\%) & SER (dB) & PSNR (dB)  \\
\midrule
    Random & 3.85      & 14.56      & 29.36       \\
 Ada\_Coh~\cite{krishnamurthy2013low}  & 1.52  & 18.56 &  33.67      \\
\textbf{Proposed} & \textbf{0.85}  & \textbf{21.22} &  \textbf{35.84}     \\
\bottomrule
\end{tabular}
\end{table}

\section{Conclusion}
\label{sec:conclusion}
In this paper, we have presented a tensor-format active sampling model for reconstructing high-dimensional low-rank MR images. 
The proposed $k$-space active sampling approach is based on the Query-by-Committee method. 
It can easily handle various pattern constraints in practical MRI scans. 
Numerical results have shown that the proposed active sampling methods outperform the existing matrix-coherence-based adaptive sampling method. 
In the future, we will extend the reconstruction to an online setting and verify it on more realistic MRI data. 




\newpage
\small{
\bibliographystyle{Bib/IEEEbib}
\bibliography{Bib/reference}
}

\end{document}